\documentclass[runningheads,a4paper]{llncs}
\usepackage{amssymb}
\usepackage{graphicx}
\usepackage{cite}
\usepackage{algorithm}
\usepackage{algorithmic}
\usepackage{booktabs}
\usepackage{amsmath}
\usepackage{url}
\urldef{\mailsa}\path{lvjiancheng@scu.edu.cn}
\newcommand{\keywords}[1]{\par\addvspace\baselineskip
\noindent\keywordname\enspace\ignorespaces#1}
\begin{document}

\mainmatter  % start of an individual contribution

\title{Learning Inverse Mapping by AutoEncoder based Generative Adversarial Nets}

\titlerunning{Learning Inverse Mapping by AEGAN}

\author{Junyu Luo \and Yong Xu \and Chenwei Tang \and Jiancheng Lv}
%


\institute{Machine Intelligence Laboratory, College of Computer Science, Sichuan University,\\ Chengdu 610065, P. R. China.\\
\mailsa}


\maketitle

\begin{abstract}
The inverse mapping of GANs'(Generative Adversarial Nets) generator has a great potential value.
Hence, some works have been developed to construct the inverse function of generator by directly learning or adversarial learning.
While the results are encouraging, the problem is highly challenging and the existing ways of training inverse models of GANs have many disadvantages, such as hard to train or poor performance.
Due to these reasons, we propose a new approach based on using inverse generator ($IG$) model as encoder and pre-trained generator ($G$) as decoder of an AutoEncoder network to train the $IG$ model. In the proposed model, the difference between the input and output, which are both the generated image of pre-trained GAN's generator, of  AutoEncoder is directly minimized. The optimizing method can overcome the difficulty in training and inverse model of an non one-to-one function.
We also applied the inverse model of GANs' generators to image searching and translation.
The experimental results prove that the proposed approach works better than the traditional approaches in image searching.
\end{abstract}
\section{Introduction}
Generative adversarial nets(GANs)\cite{Goodfellow2014Generative}, based on the minimax two-player game theory, show a great power in generating high quality artificial data.
And the method of Deep convolutional generative adversarial nets\cite{Radford2015Unsupervised} shows the great potential on the mapping between image space $X$ and latent space $Z$.
Lots of papers\cite{Dong2017Unsupervised,Perarnau2016Invertible,Radford2015Unsupervised} have shown the huge power of inverse model of a generator on semi-supervised learning and adjusting the outputs images of the generators.
In addition, finding the inverse mapping of generator can also provide us useful insights to the generator and we may use this to improve the performance of generator.
Building on ideas from these many previous works, many works\cite{Creswell2016Inverting,Donahue2016Adversarial,Dumoulin2016Adversarially,Perarnau2016Invertible} have been developed to learn the the inverse mapping of the generator.
But, the mapping from latent space $Z$ to image space $X$ is an uniderection mapping and non-linear inverse problem. This brings a  great challenge for finding the inverse mapping of generator.

Dumoulin and Donahue\cite{Donahue2016Adversarial,Dumoulin2016Adversarially} proposed a way of learning encoder network $E$ alongside the generator $G$ and discriminator $D$.
The approach successfully avoids the problem of uniderection mapping brought by directly training the inverse model.
However, the reconstruction results are not satisfying enough.
Creswell's idea\cite{Creswell2016Inverting} can get a good correlation between samples and reconstructions.
The approach takes the desired output $z$ as optimization goal.
It' very simple but slow, because we have to calculate the $z$ by using multiple gradient descents every step. In other words, the approach is trying to search the $z$ instead of calculating the $z$.
Perarnau proposed a invertible conditional GANs(ICGAN)\cite{Perarnau2016Invertible}.
In the approach, the inverse model is trained through directly minimizing the difference between $(E_{z},E_{y})$ and $(z,y)$, where $y$ is the label information of $x$ and $z$ is a noise vector.
While the strategy mitigates the effect of the problem that the function of generator is not
a one-to-one function, the freedom of latent space $Z$ is restricted by the label information. Due to the approach requires abundant label information, the requirement for data sets is very strict.

In this paper, we propose a new approach to learn the inverse mapping by AutoEncoder based on GANs(AEGAN) as a complement of former works.
We use the AutoEncoder to train a inverse model.
The pre-trained generator is regarded as the decoder part of an AutoEncoder and the inverse generator is regarded as the corresponding encoder part.
This model does not directly minimize the difference between the original noise vector and the reconstructed noise vector, but try to minimize the difference between the generated samples of noise vectors.
This strategy not only avoids problems of directly training the inverse model, but also avoids the poor correlation of adversarial training. In addition, we also explore the application value of inverse model in image processing.
The corresponding vectors of images contain rich semantic information.
Our experiments prove that such semantic information is very helpful in image searching.
And by combing the generator model, the inverse model can also be used in image-to-image translation.
\section{Primary and Motivation}
The existing ways of learning the inverse model of the GANs have made great success, however there still remains many problems waiting for solving.
The idea of Dumoulin and Donahue\cite{Donahue2016Adversarial,Dumoulin2016Adversarially} is to train encoder network $E$ alongside the generator $G$ and discriminator $D$.
The training objective is defined as a minimax objective:
\begin{equation}
\mathop{\mathrm{min}} \limits_{G,E}\mathop{\mathrm{max}} \limits_{D}V(D,E,G)
=\mathbb{E}_{x{\sim}p_{data}(x)}[\log D(x,E(x))] + \mathbb{E}_{z{\sim}p(z)}[\log (1 - D(G(z),z))]
\end{equation}
Where $D$,$E$,$G$ are Discriminator, Encoder, Generator respectively. $x$ is an input sample and $z$ is a noise vector. During the training, the $G$,$E$ try to minimize the value function and $D$ tries to maximize the value function.
This approach successfully avoids the problem of directly training the inverse model.
However, this approach also results the poor correlation between samples and reconstructions, because the discriminator only focus on the difference between data sets instead of the difference between two images.
So the encoder part can't catch the unique features of one signal image.
In addition, the approach needs to train a third network with the generative net, which means that inversion cannot be learned from a pre-trained generative network.

Creswell\cite{Creswell2016Inverting} proposes a different idea that can get a good correlation between samples and reconstructions.
The main idea is to directly minimize the difference between generated image $G(z)$ and sample image $x$ through optimizing the value of $z$, where $G$ is a pre-trained generator.
The $z$ is updated by:
\begin{equation}
z=z-\alpha\nabla_{z}[-x* \log (G(z))-(1-x)* \log (1-G(z))]
\end{equation}
Where $\alpha$ stands for the learning rate.
This approach takes the desired output $z$ as optimization goal.
It's easy to implement but poor in effectiveness because it doesn't provide a real inverse function and we have to use gradient descents every time.

The invertible conditional GAN(ICGAN)\cite{Perarnau2016Invertible}, proposed by Perarnau, tries to solve the problem through conditional GAN. In this model, ICGAN tries to minimize the difference between $(E_{z},E_{y})$ and $(z,y)$, $y$ is the label information including the gender, age and so on.
$E_{z}$ is the noise vector Encoder and $E_{y}$ is the label information Encoder.
The two training objectives are:
\begin{equation}
\begin{aligned}
&L_{ez}=\mathbb{E}_{z{\sim}p_{z},y^{'}{\sim}p_{y}}\left \|z-E_{z}(G(z,y^{'}))\right \|_{2}^{2}\\
&L_{ey}=\mathbb{E}_{x,y{\sim}p_{data}}\left \|y-E_{y}(x)\right \|_{2}^{2}
\end{aligned}
\end{equation}
The label information limits the freedom of latent space $Z$.
This strategy mitigates the effect of the uniderection mapping problem . But this approach requires abundant label information, which results that it can not be used in unsupervised approach.
\section{AutoEncoder based Generative Adversarial Nets}
The details of training and network structure can be found int he Appendix.
\subsection{Basic Structure}
Our idea of AEGAN is inspired from AutoEncoder.
Here we take the generator $G(z;\theta _{g})$ as the decoder part of the AutoEncoder and the desired inverse generator $IG(x;\theta _{ig})$ as the encoder part.
Fig.~1 shows the training process of AEGAN, we try to minimize the difference between generated image $x$ and reconstructed image $x'$.
$IG$ compresses a generated image $x$ into a latent space vector $z'$ and $G$ reconstructs the $z'$ into a new image $x'$. $z'$ is used as the extracted feature of input sample $x$ and our experiments prove that $z'$ is a very good image feature in image translation and searching. Many previous methods regard the generator as an encoder part instead of a decoder part in training and it's against the nature of AutoEncoder.
And the mapping from $z$ to $x$ also brings difficulty to the learning of the encoder part.
So in AEGAN, we change our goal into minimizing $|x-x'|$.
\newline
The innovation of AEGAN is that we focus on the reconstructed images instead of the reconstructed noise vectors.
This model does not directly minimize the difference between $z$ and $z'$, but try to minimize the difference between the $x$ and $x'$.
This strategy not only avoids problems in directly training the inverse model, but also avoids the poor correlation of adversarial training.
\begin{figure}
\begin{center}
\includegraphics[width=0.85\linewidth]{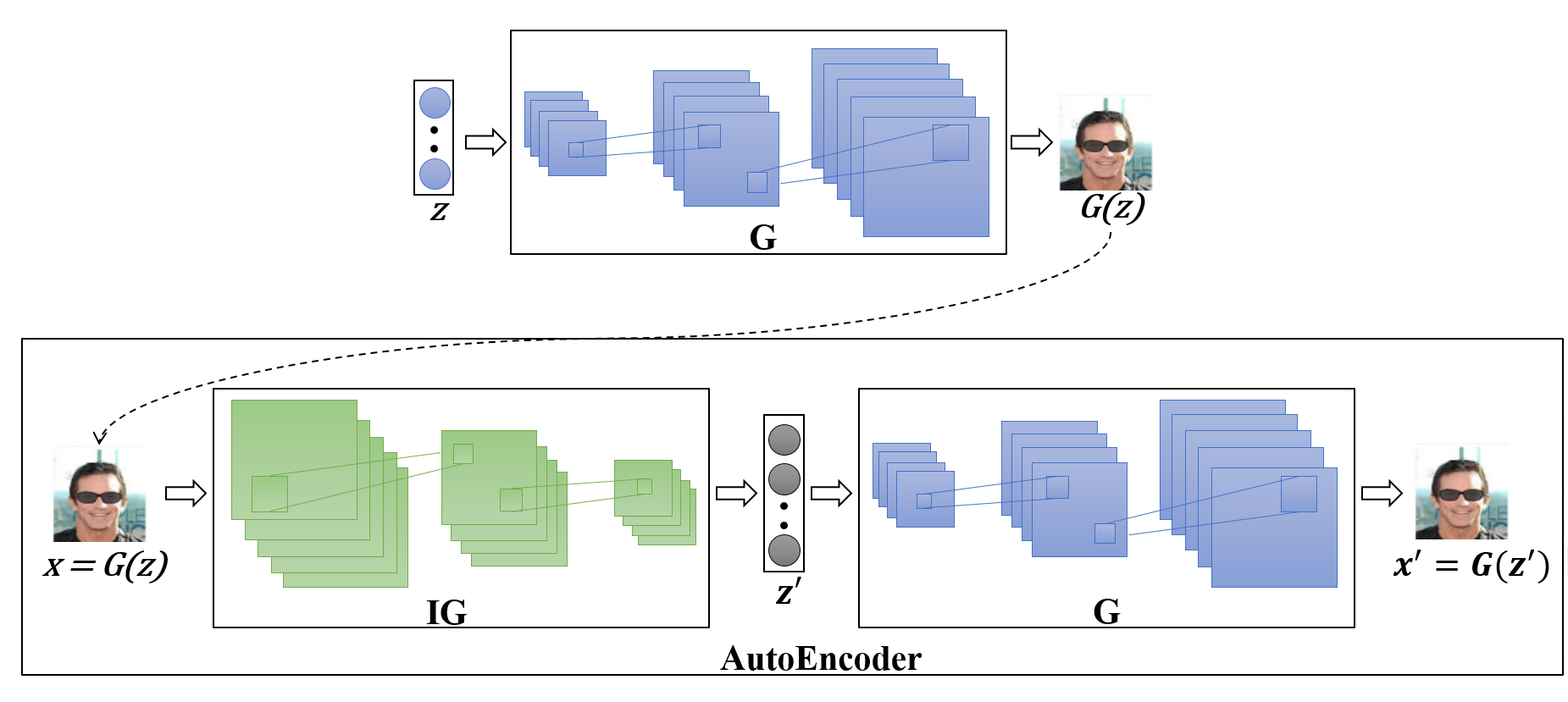}
\caption{The training process of AEGAN}
\end{center}
\label{fig:AEGAN}
\end{figure}
\subsection{Training Steps}
\subsubsection{Training the Generator}
First we train the GAN's generator $G$ using the approach and the network structure of DCGAN\cite{Radford2015Unsupervised}. $G$ is a deconvlutional network with one fully connected layer and four deconvlutional layers with strides $(1,2,2,1)$. The activation function is relu for first four layers and sigmoid for the last layer. The sigmoid layer is aimed at normalizing the generated images. And prior $z\in \mathbb{R}\sim \mathrm{U}(-1,1)$.
\newline
The optimization goal of generator is:\newline
\begin{equation}
\begin{aligned}
\mathop{\mathrm{min}} \limits_G
\mathop{\mathrm{max}} \limits_D
V(\theta _{d},\theta _{g})=\mathbb{E}_{x{\sim}p_{data}(x)}[\log D(x)] + \mathbb{E}_{z{\sim}p(z)}[\log (1 - D(G(z)))]
\end{aligned}
\end{equation}
\subsubsection{Training the Inverse Generator}
Then we start to train the inverse generator $IG$ by using the information from a pre-trained generator $G$.
In details, the structures of $G$ and $IG$ are symmetric. The deconvolutional layers of $G$ are replaced with the corresponding convolutional layers. The activation function of output layer is tanh for limiting the range of reconstructed $z'$. The convolution type in $IG$ is strided convolution \cite{Radford2015Unsupervised}.
To avoid the difficulty of directly training the encoder we require the value function of $IG$ to minimize the difference between the fake image $x$ generated by $G$ and the reconstructed output $x'$.
We choose the cross-entropy function to define the difference between $x$ adn $x'$.
The optimization objective can be defined as:
\begin{equation}
\begin{aligned}
&\mathop{\mathrm{min}}\limits_{IG}\mathbb{E}_{x{\sim}p_{generated}(x)}\{V(x;\theta _{ig})\}\\
V(x;\theta _{ig})=-x&*\log x'-(1-x)* \log (1-x')\\
=-x&*\log G(IG(x;\theta _{ig}))-(1-x)* \log (1-G(IG(x;\theta_{ig})))
\end{aligned}
\end{equation}
Where $\theta _{g},\theta _{d}$ are the parameters of the generator $G$ and discriminator $D$.
\newline
Algorithm 1 shows the detail of training $IG$.
\begin{algorithm}[H]
\caption{Training the Inverse Generator}
\label{alg:A}
\begin{algorithmic}
\FOR{number of training iterations}
\STATE 1.Sample minibatch of $m$ noise samples $(z^{(1)},...,z^{(m)})$ from noise prior $z{\sim}p_{g}(z)$ and use them to generate the training images $(x^{(1)},...,x^{(m;.;.)}){\sim}p_{generated}(x)$ through the pre-trained generator $G$.
\STATE 2.Put the generated image $x(x=G(z))$ into the AutoEncoder part to get the reconstructed image $x'$.
\begin{equation}
\begin{aligned}
z' &= IG(x) = IG(G(x))\\
x' &= G(z')
\end{aligned}
\end{equation}
\STATE 3. Compute the reconstruction loss $V(x)$ according to Equation (4).
\begin{equation}
\begin{aligned}
V(x)=-x*\log G(IG(x))-(1-x)* \log (1-G(IG(x)))
\end{aligned}
\end{equation}
\STATE 4. Perform a backpropagation to compute the gradients and only upgrade the parameters of $IG$.
\begin{equation}
\theta_{ig} = \theta_{ig} - \frac{\alpha}{m}\sum_{i=1}^{m}\frac{\partial V(x^i;\theta _{ig})}{\theta _{ig}}
\end{equation}
Where $\alpha$ is the learning rate.
\ENDFOR
\end{algorithmic}
\end{algorithm}
\section{Experiment Results}
We evaluate the ability of this inverse model on CelebFaces Attributes Dataset (CelebA)\cite{liu2015faceattributes}.
CelebA is a large-scale face attributes dataset.
\subsection{Reconstructing Samples}
We take the outputs of generator as the samples and use the inverse mapping from these samples to $Z$ space to generate the reconstructed samples.
Here, we compare AEGAN with a directly trained inverse model based on ICGAN\cite{Perarnau2016Invertible} and the adversarial inverse model based on BiGAN\cite{Donahue2016Adversarial}. The original BiGAN and ICGAN both contain an addtional image label information vector. In here we remove the label vectors because of the unsupervised condition.
Fig.~2 shows the reconstructed results of AEGAN, inverse model and BiGAN.
For BiGAN we uses the different original samples because BiGAN can't use a pre-trained generator as base. This is because that in BiGAN the generator and inverse generator are trained in the same time as Equation (1) shows.
So we compare BiGAN with the generated samples from its own generator for fairness.
In addition, we use the dHash\cite{Niu2008An} as standard to evaluate the similarity of generated images. dHash will give every image a special hash code and the difference between hash codes can be used to describe the similarity between images.
We take the average similarity as final result.
As we can see in Table 1, the result of AEGAN is also the best in this experiment.
\begin{figure}
\begin{center}
\includegraphics[width=0.8\linewidth]{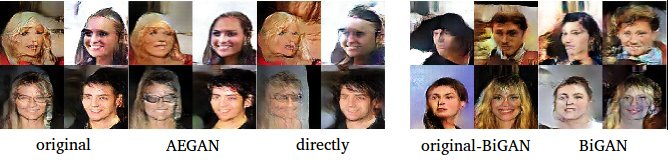}
\caption{The reconstructed results}
\end{center}
\label{fig:reconstructuion}
\end{figure}
\begin{table}
\centering
\caption{Similarity compared with original samples}
\begin{tabular}{lcl}
\hline\noalign{\smallskip}
AEGAN&Directly training&BiGAN\\
\noalign{\smallskip}
\hline
\noalign{\smallskip}
0.8266&0.7944&0.6594\\
\hline
\end{tabular}
\end{table}
\subsection{Searching the Similar Images Using AEGAN}
To illustrate the power of AEGAN, we will show its ability in searching the similar images.
We only compare with the general image searching algorithm, because our approach is based on unsupervised learning.
We compare AEGAN with three general image searching algorithms: dHash, pHash\cite{Niu2008An} and color histogram\cite{Swain1991Color}.
In details, the similarity between two images are based on the Euclidean Distance between the reconstructed $z'$ vectors of them, the smaller the distance, the higher the similarity.
We take an image from the original data set celebA and add some other
factors such as color transform, adding a sunglasses to the person, to form the 3 test images. Then we implement 4 different algorithms to find the closest images in the first 20,000 images of celebA.The Fig.~3 proves that AEGAN is very suitable for this task.
To evaluate the comprehensive performance, the second experiment is aiming at finding the similar images.
We compare our algorithm with dHash. We take the first 20,000 images of celebA as the test set and take other 64 images from the celebA as base images. As we can see from Fig.~4, AEGAN approach is much better than dHash. AEGAN catches the important features of face images such as the face angle, face similarity, hair style and facial expression.
In addition, we use the label similarity to evaluate the searching resluts. There are 40 labels for each image including gender, hair color and so on. The result can be seen from Table 2.
Although AEGAN is not a patch on specialized face recognition algorithms in this task, we have to emphasize that this approach is unsupervised and universal. In other words, this idea can easily be implemented in other fields.
\begin{table}
\centering
\caption{Label similarity compared with base samples}
\begin{tabular}{lcl}
\hline\noalign{\smallskip}
AEGAN&dHash\\
\noalign{\smallskip}
\hline
\noalign{\smallskip}
0.7918&0.7483\\
\hline
\end{tabular}
\end{table}
\begin{figure}
\label{fig:totala}
\begin{center}
\includegraphics[width=0.52\linewidth]{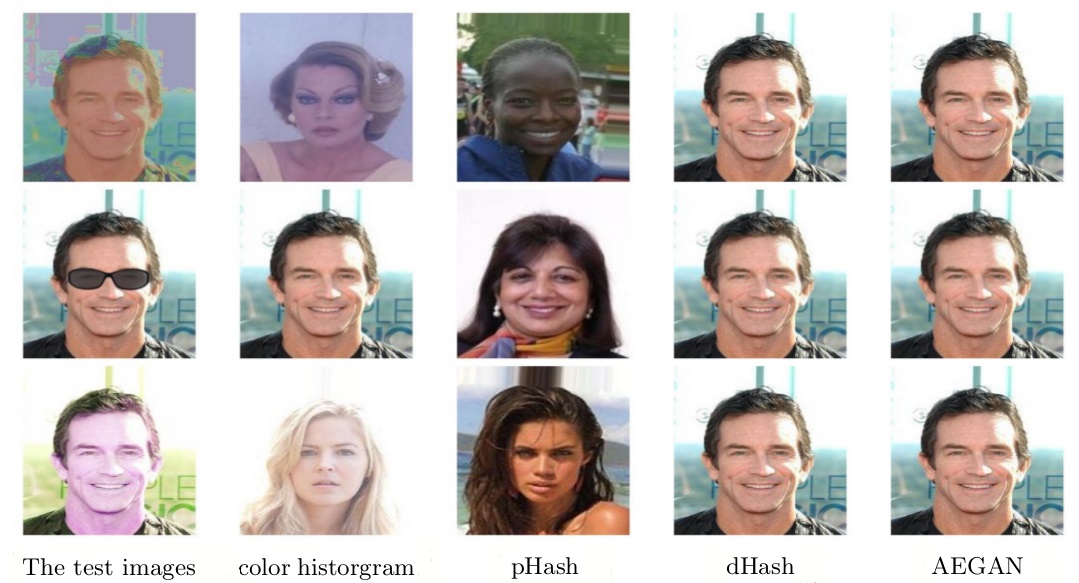}
\caption{The searching results.}
\end{center}
\end{figure}
\begin{figure}
\begin{center}
\includegraphics[width=0.75\linewidth]{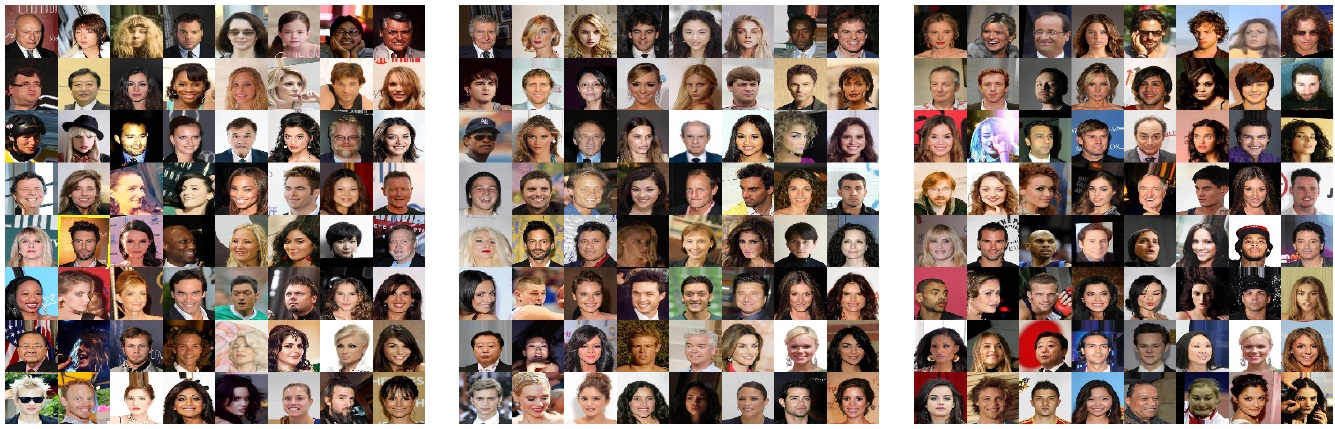}
\caption{The searched results for dHash and AEGAN. The first block contains the base images. And the second one is the searching result of AEGAN. The last one is the result of dHash.}
\end{center}
\label{fig:totalb}
\end{figure}
\subsection{Super-Resolution Using the AEGAN}
To prove our approach does learn the major features of face images, we propose the third experiment.
In this experiment we take the Gaussian Blur images as inputs of inverse generator $IG$ and then use the output of inverse generator to reconstruct the original images. We choose the generated data as the original examples. As Fig.~5 shows, we can see AEGAN also performs well in super-resolution and we didn't train the AEGAN specially for this task. AEGAN can automatically ignore the abnormal parts of input sample and add the missing features to the reconstructed output.
It is worth mentioning that the approach is unsupervised. As we known, the labeled data are extremely rare in most of application. Given the data limitations, the proposed method work surprisingly well for Super-Resolution without label information.
\begin{figure}
\begin{center}
\includegraphics[width=0.75\linewidth]{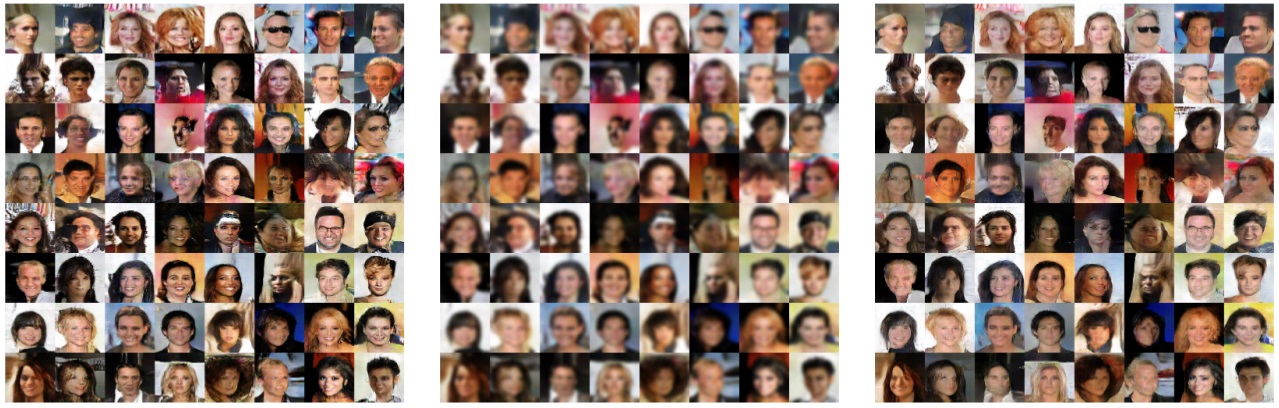}
\caption{The results of super-resolution by AEGAN. The first block contains the base image. And the second block contains the images after adding the Gaussian Blur. The last one is the reconstructed result.}
\end{center}
\label{fig:totalc}
\end{figure}
\section{Conclusion And Further Works}
AEGAN uses the idea of Auotoencoder to overcome the difficulty in training a inverse model of generator.
And the experiments show that the inverse mapping of generator has a very similar function compared with Word Embedding\cite{Hinton1986Learning}. Because the inverse out put of an image can be regarded as a vector presentation of the image and this vector presentation can catches the important features of images as Experiment 2 shows.
This ability can be very helpful in fields of image and video processing.
It's possible to get a universal vector representation of image if we train the AEGAN at large image data sets.
In addition, we can use AEGAN to reform the Image-to-Image Translation approach based on GAN\cite{Isola2016Image}.
With AEGAN, the training of generator part can be done in unsupervised condition and we only need to train the encoder part in conditional situation.
In other words, it's possible to train a Image-to-Image GAN net in semi-supervised condition if we use the structure of AEGAN.

\bibliographystyle{splncs03}

\begin{thebibliography}{4}

\bibitem{Goodfellow2014Generative} Goodfellow, I.J., Pougetabadie, J., Mirza, M., Xu, B., Wardefarley, D., Ozair, S.,
Courville, A., Bengio, Y., Ghahramani, Z., Welling, M.: Generative adversarial
nets. Advances in Neural Information Processing Systems 3, 2672--2680 (2014)

\bibitem{Radford2015Unsupervised} Radford, A., Metz, L., Chintala, S.: Unsupervised representation learning with deep
convolutional generative adversarial networks. arXiv preprint arXiv:1511.06434 (2015)

\bibitem{Dong2017Unsupervised} Dong, H., Neekhara, P., Wu, C., Guo, Y.: Unsupervised image-to-image translation
with generative adversarial networks. arXiv preprint arXiv:1701.02676 (2017)

\bibitem{Perarnau2016Invertible} Perarnau, G., van de Weijer, J., Raducanu, B., Alvarez, J.M.: Invertible conditional ´
gans for image editing. arXiv preprint arXiv:1611.06355 (2016)

\bibitem{Creswell2016Inverting} Creswell, A., Bharath, A.A.: Inverting the generator of a generative adversarial
network. arXiv preprint arXiv:1611.05644 (2016)

\bibitem{Donahue2016Adversarial} Donahue, J., Kr¨ahenb¨uhl, P., Darrell, T.: Adversarial feature learning. arXiv
preprint arXiv:1605.09782 (2016)

\bibitem{Dumoulin2016Adversarially} Dumoulin, V., Belghazi, I., Poole, B., Lamb, A., Arjovsky, M., Mastropietro, O.,
Courville, A.: Adversarially learned inference. arXiv preprint arXiv:1606.00704
(2016)

\bibitem{liu2015faceattributes} Liu, Z., Luo, P., Wang, X., Tang, X.: Deep learning face attributes in the wild.
In: Proceedings of the IEEE International Conference on Computer Vision. pp.
3730--3738 (2015)

\bibitem{Niu2008An} Niu, X.M., Jiao, Y.H.: An overview of perceptual hashing. Acta Electronica Sinica
36(7), 1405--1411 (2008)

\bibitem{Swain1991Color} Swain, M.J., Ballard, D.H.: Color indexing. International Journal of Computer Vision 7(1), 11--32 (1991)

\bibitem{Hinton1986Learning} Hinton, G.E.: Learning distributed representations of concepts. In: Proceedings of the eighth annual conference of the cognitive science society. vol. 1, p. 12. Amherst, MA (1986)

\bibitem{Isola2016Image} Isola, P., Zhu, J.Y., Zhou, T., Efros, A.A.: Image-to-image translation with conditional adversarial networks. arXiv preprint arXiv:1611.07004 (2016)

\end{thebibliography}

\section*{Appendix: Network Structures and Training Details}
The Adam optimizer is used for all the experiments and the parameters are the same. The learning rate is 0.0002 and beta1 is 0.5. The experiment is trained on complete CelebA dataset. Batch size is 64.
\begin{table}
\centering
\caption{Generator}
\begin{tabular}{llllll}
\hline\noalign{\smallskip}
Input Shape & 100 \\
Operation       & Kernel & Stride & Filter & BN & Activation \\
\hline\noalign{\smallskip}
Dense &        &        & 4*4*64*8          & N          &            \\
Reshape         &        &        & 4,4,64*8        & Y          & Relu       \\
Deconv          & 5*5    & 2*2    & 64*4              & Y          & Relu       \\
Deconv          & 5*5    & 2*2    & 64*2              & Y          & Relu       \\
Deconv          & 5*5    & 2*2    & 64*1              & Y          & Relu       \\
Deconv          & 5*5    & 2*2    & 3                 & N          & Sigmoid  	\\ 
\hline
\end{tabular}
\end{table}
\begin{table}
\centering
\caption{Discrimiator}
\begin{tabular}{llllll}
\hline\noalign{\smallskip}
Input Shape & 64*64*3 \\
Operation       & Kernel & Stride & Filter & BN & Activation \\
\hline\noalign{\smallskip}
Conv          & 5*5    & 2*2    & 64*1              & Y          & Lrelu      \\
Conv          & 5*5    & 2*2    & 64*2              & Y          & Lrelu      \\
Conv          & 5*5    & 2*2    & 64*4              & Y          & Lrelu      \\
Conv          & 5*5    & 2*2    & 64*8              & Y          & Lrelu      \\
Reshape       &        &        & 4*4*64*8          & N          &            \\
Dense 		  &        &        & 1                 & N          & Sigmoid    \\
\hline
\end{tabular}
\end{table}
\begin{table}
\centering
\caption{Inverse Generator}
\begin{tabular}{llllll}
\hline\noalign{\smallskip}
Input Shape & 64*64*3 \\
Operation       & Kernel & Stride & Filter & BN & Activation \\
\hline\noalign{\smallskip}
Conv          & 5*5    & 2*2    & 64*1              & Y          & Relu      \\
Conv          & 5*5    & 2*2    & 64*2              & Y          & Relu      \\
Conv          & 5*5    & 2*2    & 64*4              & Y          & Relu      \\
Conv          & 5*5    & 2*2    & 64*8              & Y          & Relu      \\
Reshape       &        &        & 4*4*64*8          & N          &            \\
Dense 		  &        &        & 100               & N          & Tanh    \\
\hline
\end{tabular}
\end{table}
\begin{table}
\centering
\caption{Discrimiator(BiGAN)}
\begin{tabular}{llllll}
\hline\noalign{\smallskip}
Input Shape & 64*64*3, 100 \\
Operation       & Kernel & Stride & Filter & BN & Activation \\
\hline\noalign{\smallskip}
Conv          & 5*5    & 2*2    & 64*1              & Y          & Lrelu      \\
Conv          & 5*5    & 2*2    & 64*2              & Y          & Lrelu      \\
Conv          & 5*5    & 2*2    & 64*4              & Y          & Lrelu      \\
Conv cond concat & 	   &		& 64*4+100			& N	\\							
Conv          & 5*5    & 2*2    & 64*8              & Y          & Lrelu      \\
Conv          & 5*5    & 2*2    & 64*8              & Y          & Lrelu      \\
Conv          & 5*5    & 2*2    & 64*8              & Y          & Lrelu      \\
Reshape       &        &        & 1*1*64*8          & N          &            \\
Dense 		  &        &        & 1                 & N          & Sigmoid    \\
\hline
\end{tabular}
\end{table}
\end{document}